\documentclass[10pt,twocolumn,letterpaper]{article}

\usepackage{cvpr}
\usepackage{times}
\usepackage{epsfig}
\usepackage{graphicx}
\usepackage{amsmath}
\usepackage{amssymb}
\usepackage{booktabs, makecell, tabularx}
\usepackage{multirow}
\usepackage{array}
\usepackage{algorithm}
\usepackage{algorithmic}
\usepackage{amsfonts}
\usepackage{arydshln}
\usepackage{rotating}
\usepackage[hyphens]{url}  % DO NOT CHANGE THIS
\usepackage[export]{adjustbox}
\usepackage[caption=false,font=footnotesize]{subfig}
\usepackage[breaklinks=true,bookmarks=false]{hyperref}

\cvprfinalcopy % *** Uncomment this line for the final submission

\def\cvprPaperID{4368} % *** Enter the CVPR Paper ID here

\ifcvprfinal\fi
\begin{document}

%%%%%%%%% TITLE
\title{Class-Aware Robust Adversarial Training for Object Detection}
\author{Pin-Chun~Chen\textsuperscript{1,2}\thanks{Work done during a research assistantship  at Academia Sinica.} \qquad Bo-Han~Kung\textsuperscript{2} \qquad Jun-Cheng Chen\textsuperscript{2}\\
\textsuperscript{1}Columbia University\\ \textsuperscript{2}Research Center for Information Technology Innovation, Academia Sinica\\
{\tt\small pc2939@columbia.edu;~\{bhkung,~pullpull\}@citi.sinica.edu.tw}
}
\maketitle
\thispagestyle{empty}
\begin{abstract}
Object detection is an important computer vision task with plenty of real-world applications; therefore, how to enhance its robustness against adversarial attacks has emerged as a crucial issue. However, most of the previous defense methods focused on the classification task and had few analysis in the context of the object detection task. In this work, to address the issue, we present a novel class-aware robust adversarial training paradigm for the object detection task. For a given image, the proposed approach generates an universal adversarial perturbation to simultaneously attack all the occurred objects in the image through jointly maximizing the respective loss for each object. Meanwhile, instead of normalizing the total loss with the number of objects, the proposed approach decomposes the total loss into class-wise losses and normalizes each class loss using the number of objects for the class. The adversarial training based on the class weighted loss can not only balances the influence of each class but also effectively and evenly improves the adversarial robustness of trained models for all the object classes as compared with the previous defense methods. Furthermore, with the recent development of fast adversarial training, we provide a fast version of the proposed algorithm which can be trained faster than the traditional adversarial training while keeping comparable performance. With extensive experiments on the challenging PASCAL-VOC and MS-COCO datasets, the evaluation results demonstrate that the proposed defense methods can effectively enhance the robustness of the object detection models.
\end{abstract}
\vspace{-0.3cm}
%%%%%%%%% BODY TEXT
% \begin{figure} 
% \vspace{-0.05in}
%     \begin{center}
%   \subfloat[Clean Image Result\label{fig_1a}]{%
%       \includegraphics[width=0.315\linewidth]{LaTeX/147_ori_result.jpg}}
%     \hfill
%   \subfloat[Vanilla Adversarial Attack\label{fig_1b}]{%
%       \includegraphics[width=0.315\linewidth]{LaTeX/147_inf.jpg}}
%     \hfill
% %   \subfloat[Object-wise attack\label{fig_1c}]{%
% %         \includegraphics[width=0.23\linewidth]{147_owa.jpg}}
% %     \hfill
%   \subfloat[Class-wise Attack\label{fig_1c}]{%
%         \includegraphics[width=0.315\linewidth]{LaTeX/147_CWA.jpg}}
%     \hfill
%   \caption{Detection results after attacked by different adversarial examples to the vanilla SSD model. (a) the detection result of a clean image, (b) we craft the adversarial example through the 20-step PGD optimization with the budget $\epsilon=16$ on the multi-task loss as described in equation~\eqref{eq:1}, (c) the detection result of the proposed class-wise attack. These detection examples show the adversarial examples generated by the proposed method can more evenly attack all the objects occurred in the image than (b).}
%   \label{fig:Adversarialimage}
%   \end{center}
% \vspace{-0.25in}
% \end{figure}

\begin{figure}
    \centering
    \includegraphics[width=0.99\linewidth]{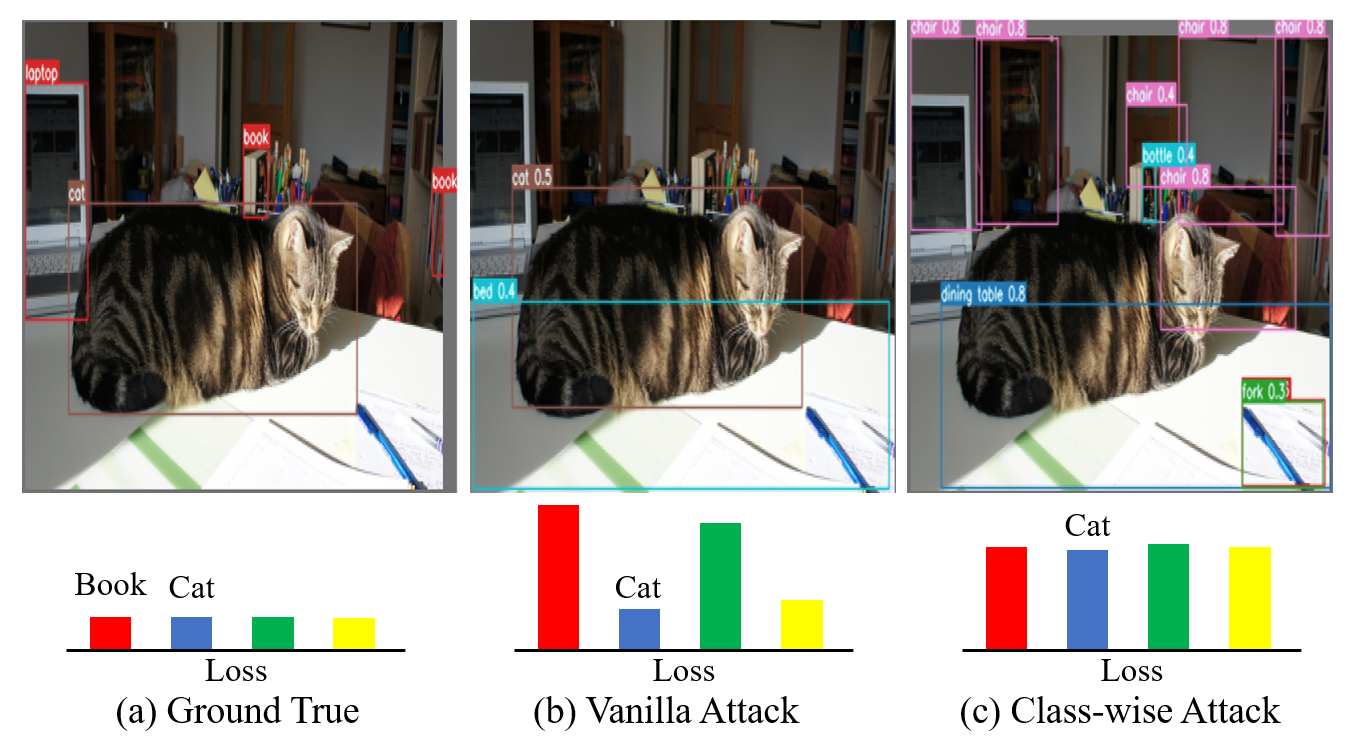}
    \caption{Detection results after attacked by different adversarial examples to the vanilla SSD model. (a) ground true, (b) we craft the adversarial example through the 10-step PGD optimization with the budget $\epsilon=8$ on the multi-task loss as described in equation~\eqref{eq:1}, (c) the detection result of the proposed class-wise attack. These detection examples show the adversarial examples generated by the proposed method can more evenly attack all the objects occurred in the image than (b).}
    \label{fig:Adversarialimage}
    \vspace{-0.5cm}
\end{figure}

\section{Introduction}\label{sec:intro}
Due to the recent breakthroughs of deep learning, deep learning-based approaches have achieved promising performance for many computer vision tasks, such as object recognition \cite{NIPS2012_4824}, \cite{DBLP:journals/corr/SimonyanZ14a}, \cite{huang-wang-2017-deep} and object detection \cite{DBLP:conf/iccv/HeGDG17}. However, researchers found there exists potential security issues for deep learning-based approaches. Szegedy et al. \cite{DBLP:journals/corr/SzegedyZSBEGF13} first crafted adversarial examples by adding imperceptible perturbations to the input images, which can easily fool the deep learning-based classification models to generate unexpected outputs. From then on, many new attack methods, including Fast Signed Gradient Method (FGSM) \cite{DBLP:journals/corr/GoodfellowSS14}, DeepFool \cite{DBLP:journals/corr/Moosavi-Dezfooli15}, Projected Gradient Descent (PGD) \cite{DBLP:conf/iclr/MadryMSTV18}, Carlini and Wagner Attack (C\&W) \cite{DBLP:journals/corr/CarliniW16a}, have been proposed to produce various adversarial examples that further expose the vulnerability of the deep learning classification models. On the other hand, object detection is one of the most important and active research fields for computer vision with plenty of real-world applications. Unfortunately, as the classification problem, it also suffers from the threat of these adversarial attacks, such as the physical adversarial patch attack to affect the steering behavior of self-driving cars \cite{DBLP:journals/corr/abs-1802-06430} or the detection results of a face detector \cite{DBLP:journals/corr/abs-1801-00349}. However, as compared with the development of attack methods, the defense algorithms to improve the robustness of object detection models are relatively few.  

In order to defend against these attacks, various methods have been proposed to enhance the robustness of the deep learning models, and one of the most effective defense approaches is adversarial training \cite{DBLP:conf/iclr/TramerKPGBM18}. In addition, for the object detection task, the approaches can be roughly categorized into two types: one-stage detector \cite{DBLP:journals/corr/LiuAESR15}, \cite{DBLP:journals/corr/abs-1804-02767} and two-stage detector \cite{DBLP:journals/corr/GirshickDDM13}, \cite{DBLP:journals/corr/Girshick15}, \cite{DBLP:conf/nips/RenHGS15}, and we focus on the one-stage detector (i.e., single-shot object detector (SSD) \cite{DBLP:journals/corr/LiuAESR15}) due to its faster detection speed and more complex nature than the two-stage detector where the nature of the two-stage detector is more similar with that of image classification task (i.e., it also performs the classification and regression tasks on the object proposals generated by the region proposal network.). Although there exists algorithms \cite{DBLP:journals/corr/abs-1907-10310} to enhance the robustness of the one-stage detector, there are still some unsolved problems: vanilla adversarial training using the overall loss of one-stage object detector does not properly take all the objects occurred in an image into consideration. As shown in equation~\eqref{eq:1}, the object detection loss of a specific object  consists of a classification loss to identify the object class and a regression loss for bounding box regression of the object. The total loss for all the occurred objects in a given image can be written as follows:
{\small \begin{equation}\label{eq:1}
\mathcal{L}= \frac{1}{N_{o}} \left ( \sum_{i=1}^{N_{o}} l_{cls}\left ( O_{i}, \left \{y_{i} \right \},\theta \right ) + l_{reg}\left (O_{i}, \left \{b_{i} \right \},\theta \right ) \right )
\end{equation}
}
where $O_i$ presents $i$-th matched default box in the image, $N_{o}$ is the number of matched default boxes, $l_{cls}$ and $l_{reg}$ are the losses of the classification branch and regression branch respectively.

As shown in Figure~\ref{fig:Adversarialimage} , not all of the detected objects in an image by an object detector can be attacked successfully if we generate the adversarial examples directly using the total loss described in the equation~\eqref{eq:1} since the sub-loss for a specific object (i.e., the loss of a specific object might go to infinity.) and a specific object class (i.e., in a given image, there are more objects of a specific class than other classes.) might dominate the overall loss value during the generation process of adversarial examples. To address these issues, we present a novel class-aware robust adversarial training for the object detection task. For a given image, the proposed approach generates an universal adversarial perturbation to simultaneously attack all the occurred objects in the image through jointly maximizing the respective loss for each object. For the classification and regression losses of each object, we clip each of them respectively to avoid the situation that the specific object loss dominates the overall loss. Meanwhile, instead of normalizing the total loss with the number of objects, the proposed approach decomposes the total loss into class-wise losses and normalizes each class loss using the number of objects for the corresponding class to mitigate the situation that the loss of a specific class dominates others. The adversarial training based on the proposed class weighted loss can not only balances the influence of each class but also effectively and evenly improves adversarial robustness of trained models for all the object classes as compared with the previous defense methods. In addition, due to the high computational cost of vanilla adversarial training, we also adopt the recent developed fast adversarial training methods~\cite{DBLP:journals/corr/abs-1904-12843} into the proposed approach to accelerate the training speed to suffice the practical needs of real-world applications. With extensive experiments on the challenging PASCAL VOC~\cite{Everingham2014ThePV} and MS-COCO~\cite{DBLP:journals/corr/LinMBHPRDZ14} datasets, the evaluation results demonstrate that the proposed defense methods can effectively enhance the robustness of the object detection models.
\begin{figure}
\begin{center}
\includegraphics[width=\linewidth]{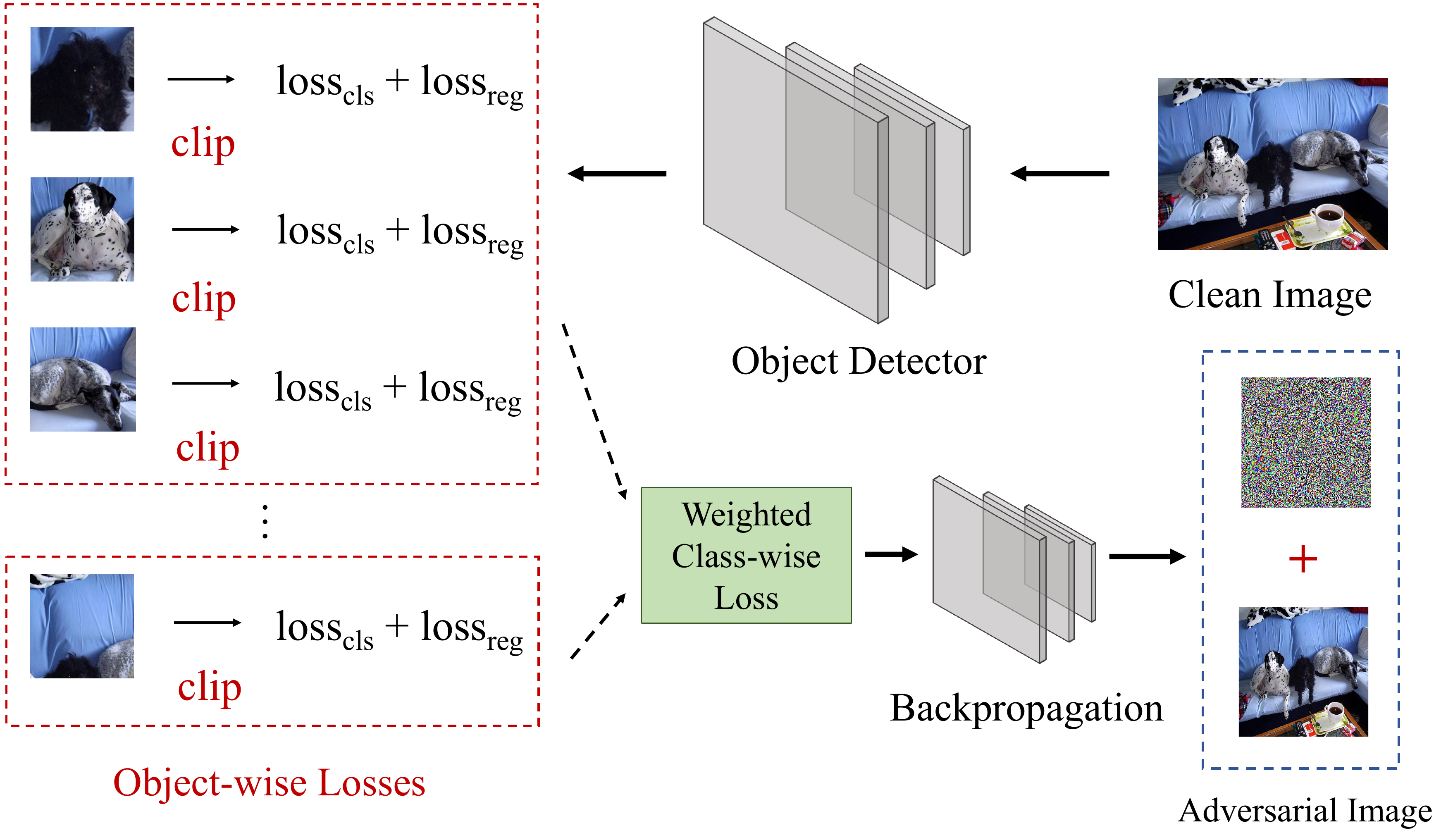}
% \vspace{-0.2in}
\caption[]{The framework of generating class-wise adversarial examples. In the process of class-wise adversarial generation, we first separate task-oriented losses into object-wise losses and clip each classification and regression loss of an object to force the values of them in the same scale. Then, we generate perturbations from the weighted class-wise loss. Finally, we add the class-wise adversarial perturbations into clean images to generate the class-wise adversarial images.}
\label{fig:flowchart}
\end{center}
\vspace{-0.2in}
\end{figure}
We summarize the main contributions of our work as follows:
\begin{itemize}
    \item We provide a systematic analysis and design several efficient and effective adversarial training algorithms for object detection, especially for the situations when there are multiple objects from different classes appearing in a given image. The proposed approaches can craft adversarial examples which can more evenly attack all the objects occurred in an image than previous methods and help improve the adversarial robustness of the trained model with adversarial training.
    \item We build the connection between the universal adversarial perturbation in the context of image classification and the object detection.
\end{itemize}

% The rest of the paper is organized as follows. Section~\ref{sec:background}, we briefly review the relevant related works. Then, we introduce the details of the proposed approach in 
% Section~\ref{sec:method}. The experimental results and discussions are presented in Section~\ref{sec:exp}. Finally, we conclude the paper in Section~\ref{sec:conclusion}.

\section{Related Works}\label{sec:background}
Due to a large amount of related works in the literature, we briefly review recent relevant works as follows.

% with special focus on (1) the adversarial attack for classification tasks, (2) its extension for the object detection tasks, and (3) recent development of adversarial training for the defense against adversarial attacks, especially about the improvement of training speed.

\subsection{Robustness of the Classification Models}
For the adversarial attacks to the deep learning-based models, Szegedy et al. \cite{DBLP:journals/corr/SzegedyZSBEGF13} first presented a method to craft adversarial examples by performing one-step back-propagation given a pretrained classification model, called Fast Gradient Sign Method (FGSM) \cite{DBLP:journals/corr/GoodfellowSS14}. FGSM emphasizes the significant impact on the direction of the gradient with respect to the input image. To defend the FGSM attack, they also proposed the adversarial training by using the adversarial images generated by performing the FGSM approach to train the original model. In this way, the worst-case error with the perturbed data would be minimized. From then on, various attacks and defense algorithms have been presented \cite{DBLP:journals/corr/KurakinGB16}, \cite{DBLP:journals/corr/abs-1808-07945} \cite{DBLP:journals/corr/Moosavi-Dezfooli15}, \cite{DBLP:journals/corr/CarliniW16a}, \cite{DBLP:journals/corr/PapernotMWJS15},
\cite{DBLP:journals/corr/abs-1712-02976},
\cite{DBLP:conf/iclr/TramerKPGBM18} for the classification tasks. Projected Gradient Descent (PGD)  is one of the strongest ``first-order adversary'' attacks as shown in \cite{DBLP:conf/iclr/MadryMSTV18} which repeats the optimization of the aforementioned saddle point formulation several times to generate adversarial examples. In addition, training using this min-max formulation makes the learned model more robust toward adversarial attacks.

Unlike ``per-instance attack'', recently Seyed-Mohsen Moosavi-Dezfooli et al. \cite{DBLP:journals/corr/Moosavi-Dezfooli16} proposed universal adversarial perturbations and Shafahi et al. \cite{DBLP:journals/corr/abs-1808-07945} proposed universal adversarial training. For these works, instead of finding per-instance perturbations for each input image, they reformulate the original optimization problem to craft a universal attack to multiple instances at a time across different classes as shown in equation~\eqref{eq:2}. Similarly, based on universal adversarial perturbation, they also proposed corresponding universal adversarial training schemes as shown in equation~\eqref{eq:3}.
{\small \begin{equation}\label{eq:2}
\underset{\delta }{\max}\;\mathcal{L}\left ( \theta, \delta  \right )=\frac{1}{N}\sum_{i=1}^{N}\textit{l}\left ( x_{i}+\delta, \left \{y_{i} \right \}, \theta  \right )\, s.t.\, \left \| \delta  \right \|_{p}\leq \epsilon 
\end{equation} 
\begin{equation}\label{eq:3}
\underset{\theta}{\min}\;  \underset{\left \| \delta  \right \|_{p}\leq \epsilon }{\max}\mathcal{L}\left ( \theta,\delta  \right )=\frac{1}{N}\sum_{i=1}^{N}\textit{l}\left ( x_{i}+\delta, \left \{y_{i} \right \}, \theta   \right )
\end{equation} }
where $\textit{l}\left (\cdot, \cdot, \theta  \right )$ represents the loss function of the training model, $\delta$ is the adversarial perturbation, and $\left \| \delta  \right \|_{p}\leq \epsilon$ denotes the $\ell_{p}$-norm to prevent $\delta$ from growing too large. Moreover, they use a ``clipped'' version of the loss function,
{\small \begin{equation}\label{eq:4}
\textit{l}\left ( x_{i}+\delta, \left \{y_{i}  \right \}, \theta   \right )=\min\left \{ \textit{l}\left ( x_{i}+\delta, \left \{y_{i}  \right \}, \theta  \right ),\beta  \right \}
\end{equation}
}
For the universal adversarial perturbation as shown in equation~\eqref{eq:4}, Shafahi et al. constrained the loss value at most $\beta$ through the clipping operations to prevent the classification loss of any single image from dominating the overall loss of multiple images as shown in equation~\eqref{eq:2}.

\subsection{Adversarial Attack for Object Detection}
% Unlike image classification problem, object detection is one of computer vision tasks that detects the occurrence of semantic objects in an image. There are two main categories: one-stage and two-stage object detectors. The detection procedure of a two-stage object detector consists of two steps: (1) the region proposal step and (2) the classification and localization step. The representative models include R-CNN \cite{DBLP:journals/corr/GirshickDDM13}, Fast R-CNN \cite{DBLP:journals/corr/Girshick15}, Mask-RCNN \cite{DBLP:conf/iccv/HeGDG17}, etc. First, they would first detect object proposals to coarsely localize the potential regions containing objects, and then classify these proposals to get the final detection results. In this paper, we focus on one-stage detector which is the essential object detection model, and the running speed of one-stage object detectors is much more efficient than the two-stage ones in solving real-world problems. YOLO \cite{DBLP:journals/corr/abs-1804-02767} and SSD \cite{DBLP:journals/corr/LiuAESR15} are two representative one-stage detectors, which simultaneously predict the bounding boxes and classify the anchors in a single inference.

Unlike image classification problem, object detection is the task to detect the occurrence of semantic objects in an image. There are two main categories: one-stage~\cite{DBLP:journals/corr/abs-1804-02767,DBLP:journals/corr/LiuAESR15} and two-stage object detectors~\cite{DBLP:journals/corr/GirshickDDM13,DBLP:journals/corr/Girshick15,DBLP:conf/iccv/HeGDG17, DBLP:conf/nips/RenHGS15}. The detection procedure of a two-stage object detector consists of two steps: (1) the region proposal step and (2) the classification and localization step.
In this paper, we focus on one-stage detectors which simultaneously predict the bounding boxes and classify the anchors in a single inference, and the running speed is much faster than the two-stage ones in solving real-world problems. 

Recently, many adversarial attacks are developed for object detection models, and most of them focus on attacking the two-stage detectors. The first attack algorithm is DAG proposed by Xie et al. \cite{DBLP:journals/corr/XieWZZXY17}, which specifies the adversarial labels and use back-propagation to iteratively mislead the predictions of the object detectors. Then, Li et al. \cite{DBLP:journals/corr/abs-1809-05962} designed RAP algorithm which combines the label loss and shape loss to yield the adversarial perturbation and optimizes the objective function with an iterative gradient based method. Wei et al. \cite{DBLP:conf/ijcai/WeiLCC19} claimed these attack methods focus on attacking the object proposal-based detector have two limitations including weak transferability and high computation cost. Therefore, they proposed UEA to generate adversarial examples using Generative Adversarial Network (GAN) framework and combine with high-level class loss and low-level feature loss.

Although many adversarial attacks have been developed for object detection during the past few years, the defense methods for object detection are rare. Zhang et al. \cite{DBLP:journals/corr/abs-1907-10310} proposed an adversarial training-based algorithm to enhance the robustness of the one-stage detector. Their algorithm decomposes the adversarial training into two task-oriented domains: $S_{cls}$ for the classification branch and $S_{reg}$ for the regression branch of the object detection loss:
{\small \begin{equation}\label{eq:5}
S_{cls} \triangleq \left \{ {x}'_{cls}\mid arg \max_{{x}'_{cls}\in S_{x}}\: l_{cls}\left ( {x}'_{cls},\left \{ y_{k} \right \}, \theta  \right ) \right \}
\end{equation}
\begin{equation}\label{eq:6}
S_{reg} \triangleq \left \{ {x}'_{reg}\mid arg \max_{{x}'_{reg}\in S_{x}}\: l_{reg}\left ( {x}'_{reg},\left \{ b_{k} \right \}, \theta  \right ) \right \}
\end{equation} }
where ${x}'_{cls}$ and ${x}'_{reg}$ represent the adversarial examples generated from each task, $S_{x}$ is defined as $S_{x}= \left \{ z\mid z\in \mathcal{B}\left ( x,\epsilon  \right )\cap \left [ 0,255 \right ]^{n} \right \}$ and $\mathcal{B}\left ( x,\epsilon  \right )= \left \{ z\mid \left \| z-x \right \|_{\infty }\leq \epsilon \right \}$ denotes the $\ell_{\infty}$-ball with the center as the clean image $x$ and the radius is the perturbation budget $\epsilon$.

Thus, they presented an adversarial training approach according to the task-oriented domain constraint $S_{cls}\cup S_{reg}$ that generates adversarial examples respectively from the object classification and bounding box regression tasks and selects the one which maximizes the overall object detection loss as shown in equation~\eqref{eq:7}.
{\small \begin{equation}\label{eq:7}
\min_{\theta}\left [ \max_{{x}'\in S_{cls}\cup S_{reg}}l\left ( {x}',\left \{ y_{k},b_{k} \right \}, \theta \right ) \right ]
\end{equation} }
where ${x}'$ is the adversarial example $x+\delta$ generated from the clean image $x$. Since the classification and regression losses are considered independently, the generated adversarial examples may not be able to effectively attack both branches of the object detector.

\section{Methodology}\label{sec:method}

To generate the adversarial examples which can effectively and evenly fool the object detector to change the detection results of all the occurred objects in an image, we develop a novel class-aware robust adversarial training for the object detection task. The proposed approach considers \emph{heterogeneous} (classification and regression tasks), \emph{multiple} (multiple objects), and \emph{balanced} (multiple classes) class losses to generate adversarial examples for a robust object detector. The overview of the proposed framework for generating class-wise adversarial examples is shown in Figure~\ref{fig:flowchart} and the details of the proposed approach are described as follows:

\subsection{Multi-task Adversarial Training for Object Detection}
To address the issues of adversarial training using the overall object detection loss, we delve into the details of the object detection loss. Different from the classification task which only contain a single loss (e.g., cross entropy loss for the classification task) to predict the results, the loss of the one-stage detector consists of two different kinds of losses: (1) the classification loss for predicting the category scores and (2) the regression loss (e.g., smooth $L_1$ loss) for predicting the box offsets from the input images to detect objects. Therefore, we have \emph{heterogeneous} sources of losses from the classification task and regression task for the generation of adversarial examples and adversarial training. We can define the following optimization problem for building the multi-task adversarial training for object detection.
{\small \begin{equation}\label{eq:8}
\begin{split}
\min_{\theta}\;  \max_{\left \| \delta  \right \|_{p}\leq \epsilon }\mathcal{L}\left ( \theta,\delta  \right ) &= \hat{l}_{cls}\left ( x+\delta, \left \{y \right \},\theta \right ) \\&+ \hat{l}_{reg}\left (x+\delta, \left \{b \right \},\theta \right )
\end{split}
\end{equation} }
where $\hat{l}_{t \in \left \{ cls, reg \right \}}\left ( \cdot ,\left \{ y, b \right \} ,\theta  \right )$ represents the confidence loss and localization loss used for the one-stage detectors respectively. Furthermore, the naive loss function as shown in equation~\eqref{eq:1} suffers from a significant impact that each task-oriented loss is unbounded, and one of the task-oriented losses can be extremely large during the adversarial training process. In the worst case, the value of the loss might go to infinity and dominate the overall object detection loss. To address this issue, we propose a ``clipped'' version for each task-oriented loss function,
{\small \begin{equation}\label{eq:9}
\hat{l}_{t \in \left \{ cls, reg \right \}}\left ( x+\delta, \left \{y \right \},\theta \right )
=  \min \left \{ \hat{l}_{t}\left ( x+\delta, \left \{y \right \},\theta \right ), \beta_{t} \right \}
\end{equation} }
As we have shown in equation~\eqref{eq:9}, this method can not only avoid the mutual interference between each task but also prevent any task-oriented loss from dominating the overall objective function for object detection by regularizing each task-oriented loss values at most $\beta_{t}$. 
%In Section~\ref{sec:exp}
In the experimental result section, we will perform an ablation studies to show the effect of clipping the loss with different thresholds, $\beta_{t}$. With this objective function, it will search an adversarial perturbation which can jointly and effectively maximizes both task-oriented losses for all the objects occurred in an image.

\subsection{Object-wise Adversarial Training for Object Detection}
Besides the task-oriented losses, we propose the second objective function that further delves into the object detection loss in the \emph{multiple} object aspect by considering the scenarios that there usually exist multiple objects in an image for the object detection task. We thus propose the object-wise adversarial training for object detection and formulate this problem as a min-max optimization problem as follows:
{\small \begin{equation}\label{eq:10}
\begin{split}
\min_{\theta}\;  \max_{\left \| \delta  \right \|_{p}\leq \epsilon }\mathcal{L}\left ( \theta,\delta  \right ) &= \sum_{i=1}^{N_{o}} \hat{l}^{o}_{cls}\left ( O_{i}+\delta, \left \{y_{i} \right \},\theta \right ) \\&+ \sum_{i=1}^{N_{o}} \hat{l}^{o}_{reg}\left (O_{i}+\delta, \left \{b_{i} \right \},\theta \right )
\end{split}
\end{equation} }
As we discuss in the multi-task adversarial training, the clipped version of the loss function would prevent the single task-oriented loss function to dominate the overall training loss. Since each task-oriented loss for a specific object (i.e., the loss of a specific object goes to infinity.) might dominate the overall loss value during the generation process of adversarial examples, we similarly propose a ``clipped'' version for the proposed object-wise loss function as shown in equation~\eqref{eq:11} which can bound the task-oriented losses of each object
at most $\beta_{o}$. We can then generate the adversarial perturbation for the multiple objects in an image using a similar way as the universal adversarial attack proposed in \cite{DBLP:journals/corr/abs-1808-07945} for the attacks of multiple images in the classification setting.
{\small\begin{equation}\label{eq:11}
\hat{l}_{t \in \left \{ cls, reg \right \}}^{o}\left ( O_{i}+\delta, \left \{y_{i} \right \},\theta \right ) =  \min \left \{ \hat{l}_{t}\left ( O_{i}+\delta, \left \{b_{i} \right \},\theta \right ), \beta_{o} \right \}
\end{equation} }
\vspace{-20pt}
\subsection{Class-wise Adversarial Training for Object Detection}
Furthermore, for the object detection task, it usually not only contains multiple objects but various classes for a given image. To prevent the loss of a specific object class from dominating the overall object detection loss (i.e., in a given image, there are more objects of a specific class than other classes.), we propose the third objective function as shown in equation~\eqref{eq:12}. Instead of normalizing the total loss with the number of objects, the proposed approach decomposes the total loss into class-wise task-oriented losses and normalizes each class loss using the number of objects for the corresponding class in a given image. The adversarial training based on the proposed class weighted loss can effectively balances the influence of each class during adversarial training.
{\small \begin{equation}\label{eq:12}
{\mathcal{L_{C}}}'= \frac{1}{C} \sum_{i=1}^{C} \frac{1}{n_{i}} \sum_{j=1}^{n_{i}} \hat{l}_{cls}^{o}\left ( O_{j}, \left \{y_{j} \right \},\theta \right ) + \hat{l}_{reg}^{o}\left (O_{j}, \left \{b_{j} \right \},\theta \right )
\end{equation} }
where $C$ is the number of classes in one image, $n_{c}$ is the number of the matched default boxes in the class $c$.

\subsection{Fast Adversarial Training for Object Detection}
Since the proposed defense methods focus on the complex object detection models, it requires much more computational resources to train a model than those of an image classification model, and the efficiency of the adversarial training is a critical point to train a robust detector within a reasonable amount of time and computational resources for the real world applications. However, the high computational cost of the iterative gradient back-propagation of the PGD-based adversarial training makes it less practical. Shafahi et al. \cite{DBLP:journals/corr/abs-1904-12843} recently proposed a fast adversarial training algorithm which recycles the gradient information to reduce the cost of generating adversarial examples when updating the model parameters. With the fast adversarial training, the training process could be 7 to 30 times faster than the original adversarial training. We thus adopt it into the proposed algorithm. Finally, the final version of the proposed algorithm takes the  \emph{heterogeneous} tasks, \emph{multiple} objects, \emph{balanced} class losses and fast training into consideration, and the details of the proposed fast class-wise adversarial training for object detection are summarized in Algorithm~\ref{alg:1}. Due to limited space, we also refer the readers to Appendix for more running time analysis for the fast adversarial training.
\begin{algorithm}
 \caption{Fast Class-wise Adversarial Training}
 \label{alg:1}
 \begin{algorithmic}[1]
 \REQUIRE dataset $D$, training epoch $N_{ep}$, perturbation bound $\epsilon$, learning rate $\gamma$
  \FOR {epoch = $1,...,N_{ep}/m$}
  \FOR {minibatch $B\sim D$}
  \FOR {$iter = 1$ to $m$}
  \STATE Compute gradient of loss with respect to $\delta$
  \STATE $d_{\delta} \leftarrow \mathbb{E}_{x\in B}\left [ \nabla _{\delta}{\mathcal{L_{C}}}' \left ( \theta,x+\delta  \right ) \right ]$
  \STATE Update $\theta$ with momentum stochastic gradient
  \STATE $g_{\theta}\leftarrow \mu g_{\theta}-\mathbb{E}_{x\in B}\left [ \nabla _{\theta}\mathcal{L} \left ( \theta,x+\delta  \right ) \right ]$
  \STATE $\theta\leftarrow \theta+\gamma g_{\theta}$
  \STATE Update perturbation $\delta$ with gradient
  \STATE $\delta \leftarrow \delta +\epsilon sign\left ( d_{\delta} \right )$
  \STATE Project $\delta$ to $\ell_{p}$-ball
  \ENDFOR
  \ENDFOR
  \ENDFOR
 \end{algorithmic} 
\end{algorithm}
\vspace{-0.1in}
\section{Experimental Results}\label{sec:exp}
In this section, we first describe the experimental settings and then show the evaluation results of the proposed adversarial defense approaches for the object detectors on the challenging PASCAL VOC~\cite{Everingham2014ThePV} and MS-COCO~\cite{DBLP:journals/corr/LinMBHPRDZ14} datasets in the following sections. 

\subsection{Datasets and Evaluation Settings}
For the PASCAL VOC dataset, we adopt the standard ``07+12'' protocol, which contains a total of $16,551$ images, $40,058$ objects, and $20$ classes for training. On the testing phase, we use the test set of the PASCAL VOC 2007 dataset with a total of $4,952$ testing images. To evaluate the performance of the object detector after adversarial attacks, we compute the average precision (AP) for the category of interest and the ``mean average precision'' (mAP) for the overall performance. For the MS-COCO dataset, we train the model using its training set in 2017 with a total of $118,287$ images, and the number of object categories is $80$. For testing, we evaluate the results using its validation set in 2017 with a total of $5,000$ images. The mAP with IoU threshold 0.5 is used for evaluating the robustness of a detector.

In addition, we also introduce the algorithms for comparison in all the experiments as follows:
% %
% Before discussing the evaluation results, we first introduce the various baselines and the proposed algorithms for object detectors used in all the experiments as follows:
% %
\begin{itemize}
\item {\bf \textit{STD}}: the object detector trained with natural training using clean images.

\item {\bf \textit{CLS}\footnote{Since the official implementation is not available, we re-implement it with the same setting as presented in \cite{DBLP:journals/corr/abs-1907-10310}. In addition, since the paper did not describe the parameter, the number of steps for the PGD-based adversarial training, we can only choose the one closest to their results as the foundation of performance comparison, and we also present the original result of CLS, REG, and MTD.}}: the model trained using $A_{cls}$ for PGD-based adversarial training.

\item {\bf \textit{REG}{{\textsuperscript{1}}}}: the model trained using $A_{reg}$ for PGD-based adversarial training.

\item {\bf \textit{MTD}{\textsuperscript{1}}/\textit{MTD-fast}}: the model that we trained with our own implementation with normal/fast adversarial training in~\cite{DBLP:journals/corr/abs-1907-10310} where we denote the generated adversarial examples as multi-task domain attack (MDA). %cite ICCV paper and the results
%paste the rebuttal 

\item {\bf \textit{TOAT}}: the model trained with the proposed task-oriented PGD-based adversarial training.

\item {\bf \textit{OWAT}}: the model trained with the proposed object-wise PGD-based adversarial training  where we also denote the generated adversarial examples as the object-wise attack (OWA).

\item {\bf \textit{CWAT}}: the model trained with the proposed class-wise PGD-based adversarial training where we also denote the generated adversarial examples as the class-wise attack (CWA).

% \item {\bf \textit{MIX-CWAT}}: the model trained with the proposed class-wise PGD-based adversarial training  with adaptive thresholds. (\emph{i.e.,} For images with a single object or a single class of objects, we want it to perform similarly to TOAT without any loss value clipping, and thus we want to set a large $\beta$ and the same thresholds as CWAT for the rest of images. For the implementation purpose, we set a single threshold for a batch of training sample. Thus, when the ratio of images with a single object or a single class of objects greater than 0.5, we use TOAT-$inf$, and $\beta=6$ otherwise.)
% 

\end{itemize}

\subsection{Implementation Details}
We conduct experiments using the one-stage detectors, SSD, with the VGG-16 network backbone as the main evaluation test bed where we use a modified version of the VGG-16 with batch normalization layers. All the models used in the experiments are fine-tuned from the pretrained SSD model using the training set of the corresponding object detection benchmarks and the SGD optimizer with an initial learning rate, $10^{-2}$, momentum, $0.9$, and weight decay $0.0005$ with the multi-box loss. The learning rates are decayed at $16th$ and $20th$ epochs respectively with the decay factor equal to $0.1$. The resolution of the resized input image is $300\times 300$. The range of the pixel values is between [0, 255] and then shifted according to the mean of pixel intensities of the whole dataset. For adversarial training, we use the budget $\epsilon = 8$ to generate the adversarial examples as the inputs. To be more specific, we denote $A_{cls}$ as the classification-task adversarial examples generated only considering the overall classification loss and $A_{reg}$ as the regression-task adversarial examples generated only considering the overall regression loss. For our fast PGD-based adversarial training, we set $m=4$. In addition, the SSD and its adversarial robust version are also trained with the online hard example mining strategy (OHEM)~\cite{shrivastava2016training} to sample hard negative samples. We set $\beta_t=\beta_o=6$ for all the experiments in this paper.

\begin{figure} 
  \vspace{-0.15in}
    \begin{center}
  \subfloat[$A_{cls}$ PGD attack\label{PGD_3a}]{%
       \includegraphics[width=0.4\linewidth]{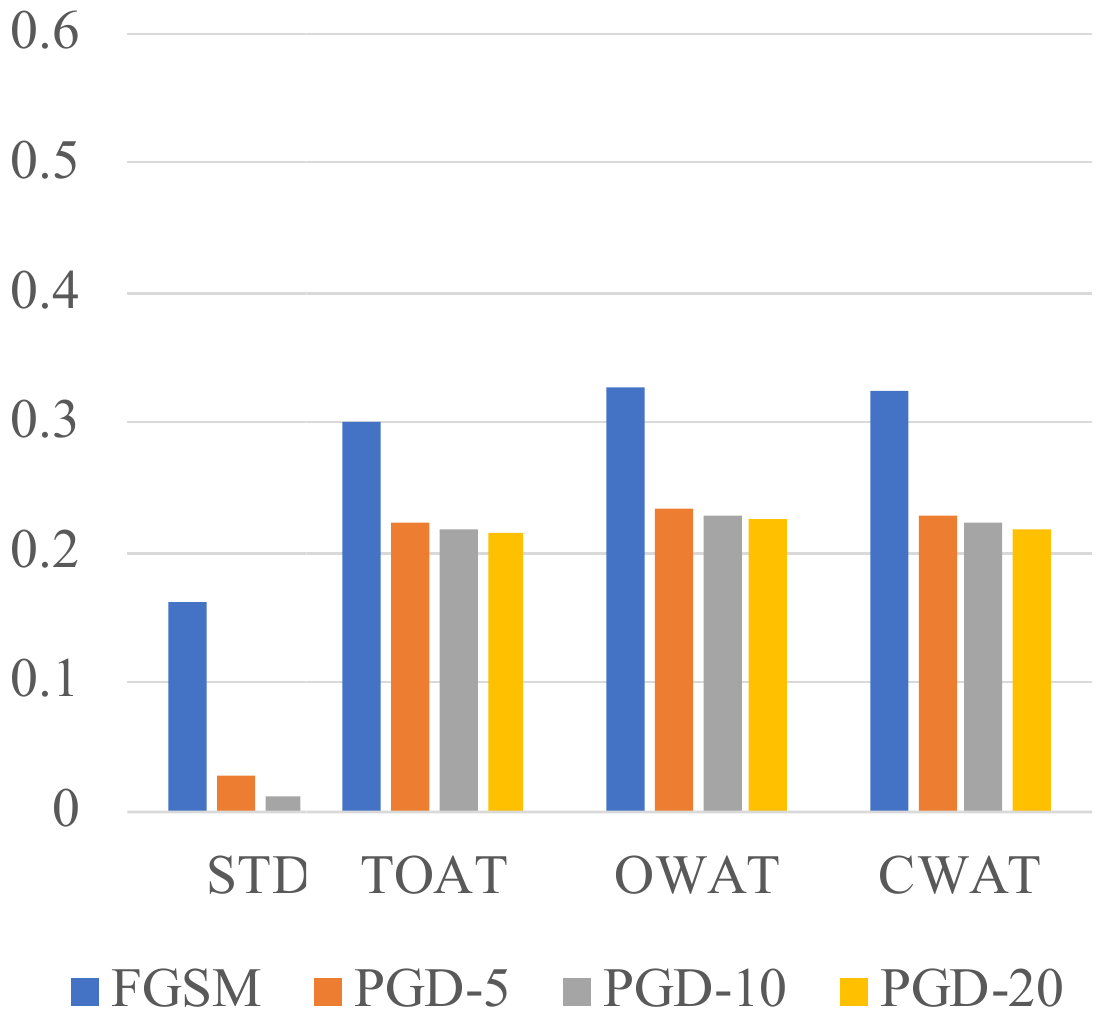}}
    % \hfill
  \quad
  \subfloat[$A_{reg}$ PGD attack\label{PGD_3b}]{%
        \includegraphics[width=0.437\linewidth]{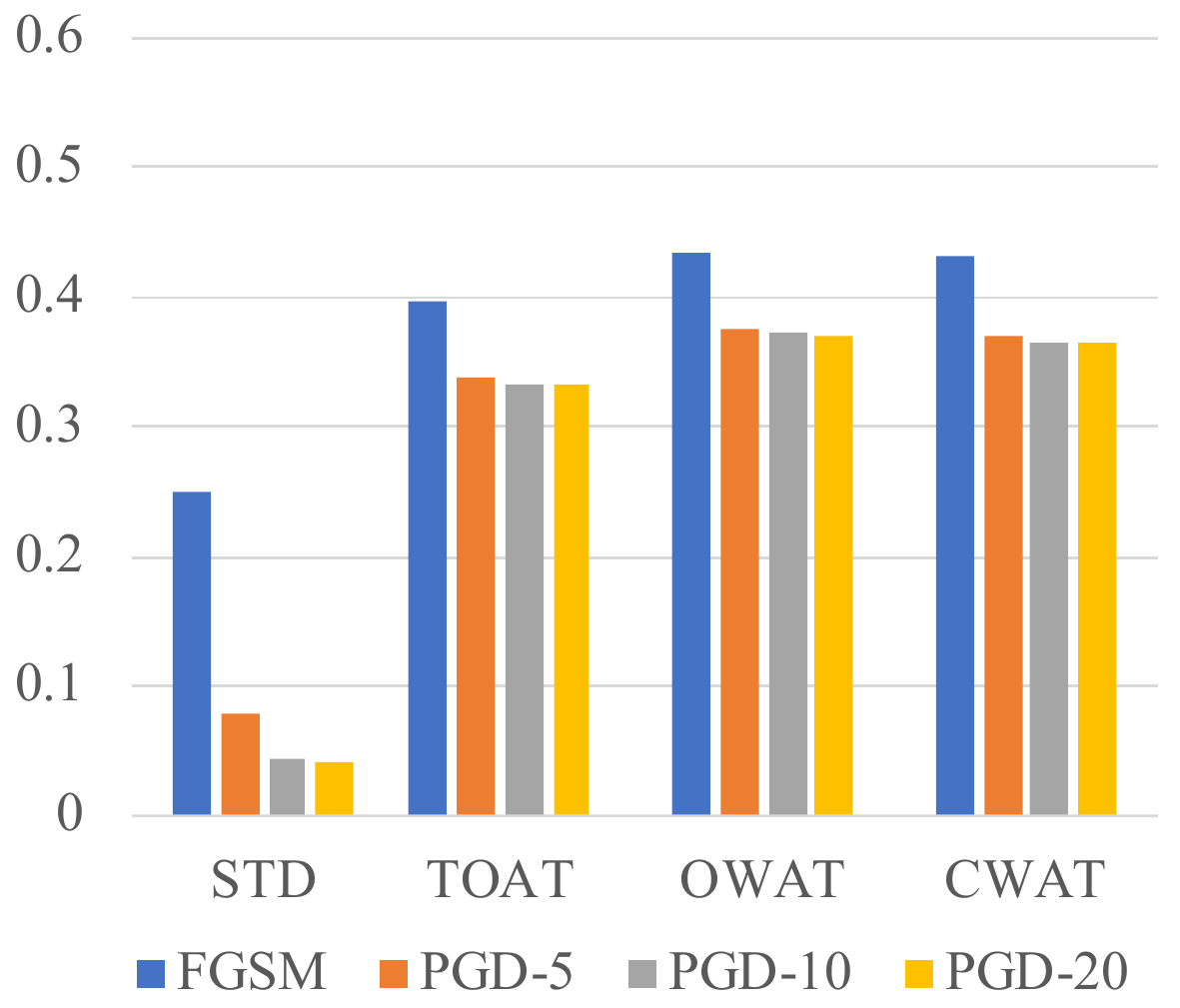}}
    \hfill
  \caption{The robustness of each model under attacks with $\epsilon=8$ from different number of steps in PASCAL VOC 2007 test set.}
  \vspace{-0.2in}
  \label{fig:diffstep}
  \end{center}
\end{figure}
%
%
% \begin{figure} 
%     \begin{center}
%   \subfloat[$A_{cls}$ PGD-10 attack\label{PGD_10_3a}]{%
%       \includegraphics[width=0.5\linewidth]{charteps_cls.pdf}}
%     \hfill
%   \subfloat[$A_{reg}$ PGD-10 attack\label{PGD_10_3b}]{%
%         \includegraphics[width=0.5\linewidth]{charteps_reg.pdf}}
%     \hfill
%   \caption{The robustness of each model under PGD-10 attacks from different budgets in PASCAL VOC 2007 test set.}
%   \label{fig:difepsmodel}
%   \end{center}
% \end{figure}
%

\subsection{Evaluation Results on Pascal VOC and MS-COCO}
\begin{figure*} 
    \begin{center}
  \subfloat[PASCAL VOC\label{o:1a}]{%
      \includegraphics[width=1\linewidth]{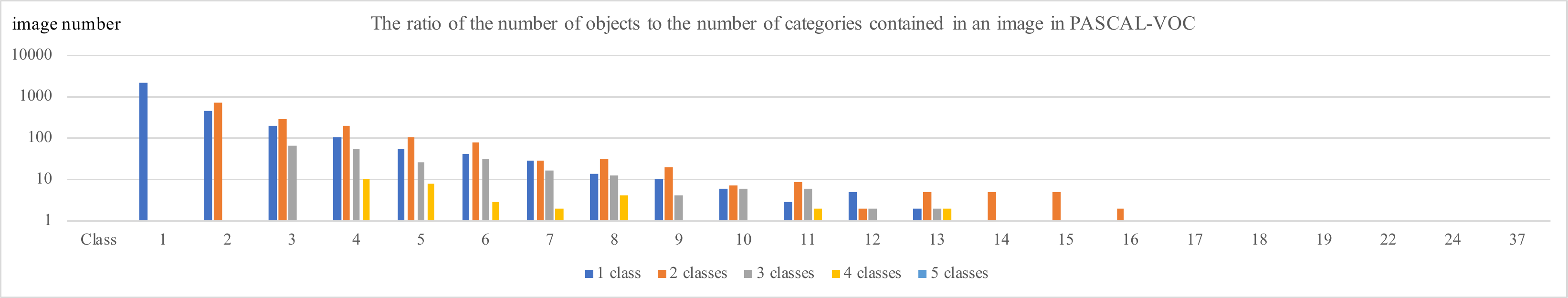}}
      \hfill\\
      \vspace{-0.3cm}
  \subfloat[MS-COCO\label{o:1b}]{%
      \includegraphics[width=1\linewidth]{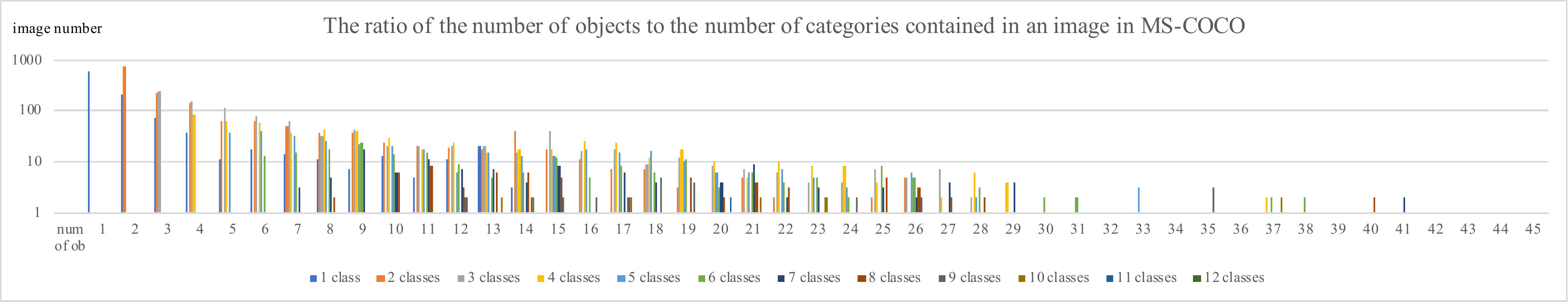}}
      \hfill
  \caption{The accumulation of number of objects and number of categories contained in an image, and the total results of the number of images toward different compositions.}
  \label{fig:objectvsclass}
  \end{center}
  \vspace{-0.5cm}
\end{figure*}

In this subsection, we show the evaluation results on both the Pascal VOC and MS-COCO datasets, where the MS-COCO dataset is more close to the real-world object detection scenarios and more challenging for testing on the robustness of object detectors. The results of different models under the PGD-10 attack with attack budget $\epsilon=8$, and other attacks, including  FGSM, CWA and DAG\footnote{We implement DAG algorithm into our one-stage models, the results are similar with the UEA experiments.}, are summarized in Table~\ref{table:summaryforfree} and ~\ref{table:cocoforfree}. 
We can find that for the Pascal VOC dataset, the proposed OWAT achieves the best performance than other compared methods while CWAT achieves comparable performance as compared with OWAT. For the MS-COCO dataset, CWAT  achieves the highest performance under the PGD-10 adversarial attack over MTD-fast, TOAT, and OWAT. As shown in Figure~\ref{fig:objectvsclass}, most of the images in the Pascal VOC contains much fewer objects than the MS-COCO, especially in different classes. Note that CWAT is less effective when the number of classes in a single image is few. That is why CWAT just achieves comparable performance to OWAT. The evaluations result of the MS-COCO dataset confirms the CWAT could effectively balance class loss to prevent the loss of a specific class from dominating the overall object detection loss, especially when we focus on comparing the results of CWAT and OWAT. 

% 
% Therefore, we independently analyze the CWAT performance with the subset that contains more than one class in an image.
% The results are illustrated in Fig.
% It can be seen that CWAT outperforms MTD, especially for multi-class images.

% \begin{figure}
%     \centering
%     \includegraphics[width=0.95\linewidth]{LaTeX/exp.png}
%     \caption{The accuracy of MTD and CWAT with regression ($A_{reg}$) and classification ($A_{cls}$) attack.}
%     \label{fig:improvement}
% \end{figure}
% we explore the impact of task-oriented losses and adversarial perturbation for the multi-object scenario in enhancing the model robustness.

%
% Then, we present a detailed ablation studies for the adversarial robustness of aforementioned robust detector trained using different loss objective under various adversarial attacks in different parameter settings for object detection as follows:

%\subsection{Evaluation Results on MS-COCO}

\subsection{Ablation Study}
% We will present a comprehensive study about the impact of \emph{heterogeneous} sources, \emph{multiple} objects and \emph{balanced} classes problem by evaluating the performance of these models under PGD adversarial attacks by task-oriented adversarial attack with different number of steps and budgets.

\begin{figure}
\vspace{-0.1in}
    \begin{center}
  \subfloat[$\epsilon=0$\label{1a}]{%
       \includegraphics[width=0.315\linewidth]{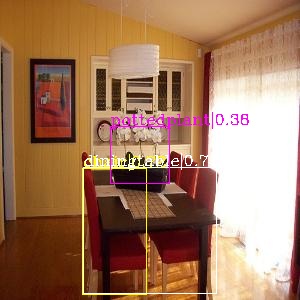}}
       \hfill
%   \subfloat[$\epsilon=2$\label{1b}]{%
%       \includegraphics[width=0.18\linewidth]{adv-2.jpg}}
%       \hfill
  \subfloat[$\epsilon=4$\label{1c}]{%
       \includegraphics[width=0.315\linewidth]{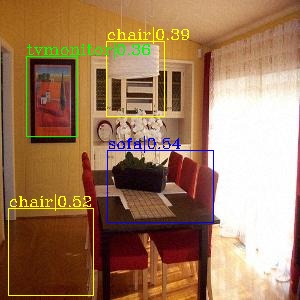}}
       \hfill
%   \subfloat[$\epsilon=6$\label{1d}]{%
%       \includegraphics[width=0.18\linewidth]{adv-6.jpg}}
%       \hfill
  \subfloat[$\epsilon=8$\label{1e}]{%
       \includegraphics[width=0.315\linewidth]{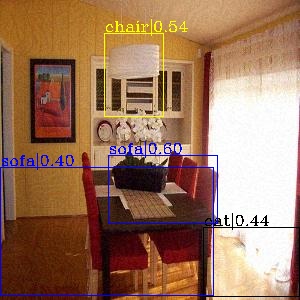}}
  \caption{The detection results of the class-wise adversarial attack with PGD-10 in different $\epsilon$, using the clean SSD as the targeted model. White label, yellow label, magenta label, blue label, green label, and black label represent classes of the dining table, chair, potted plant, sofa, tv-monitor, and respectively. (The more detailed qualitative and quantitative results can be found in Appendix.)}
  \label{fig:diffeps}
  \end{center}
  \vspace{-0.5cm}
\end{figure}
\begin{table}
% \resizebox{1\columnwidth}{!}{%
% \begin{center}
\scalebox{0.75}{
\begin{tabular}{lcp{0.05\textwidth}p{0.05\textwidth}p{0.05\textwidth}p{0.05\textwidth}cc}
\toprule
           \multirow{2}{*}{\textbf{attack}} & \multirow{2}{*}{\textbf{clean}}   & 
         \multicolumn{2}{c}{FGSM}   &
         \multicolumn{2}{c}{PGD-10} & 
         \multirow{2}{*}{\textbf{CWA}} &
         \multirow{2}{*}{DAG} \\\cmidrule(lr){3-4}\cmidrule(lr){5-6}
         &&\textbf{$A_{cls}$}&\textbf{$A_{reg}$}&\textbf{$A_{cls}$}&\textbf{$A_{reg}$}\\
\midrule
\textbf{STD} & \bf{0.752} & 0.162 & 0.25 & 0.012 & 0.043 & 0.006 & 0.291\\
%\midrule
\textbf{CLS} & 0.467 & 0.309 & 0.343 & 0.236 & 0.334 & - & -\\
%\midrule
\textbf{REG} & 0.519 & 0.263 & 0.344 & 0.146 & 0.279 & - & -\\
%\midrule
\textbf{MTD-fast} & 0.466 & 0.311 & 0.418 & 0.221 & 0.351 & 0.182 & 0.486\\
\midrule
%\textbf{TOAT-inf} & 0.469 & 0.266 & 0.374 & 0.173 & 0.310 & 0.165 & 0.456\\
%\textbf{TOAT-2} & 0.535 & 0.302 & 0.408 & 0.191 & 0.297 & 0.132 & 0.502\\
%\textbf{TOAT-4} & 0.477 & 0.288 & 0.396 & 0.190 & 0.327 & 0.183 & 0.463\\
\textbf{TOAT-6} & 0.430 & 0.300 & 0.397 & 0.218 & 0.334 & 0.197 & 0.466\\
% \textbf{TOAT-8} & 0.343 & 0.289 & 0.389 & 0.204 & 0.331 & 0.189 & 0.457\\
% \textbf{TOAT-2-4} & 0.512 & 0.304 & 0.388 & 0.211 & 0.292 & 0.131  & 0.487\\
% \textbf{TOAT-4-2} & 0.479 & 0.296 & 0.404 & 0.203 & 0.337 & 0.178 & 0.472\\
% \textbf{TOAT-2-6} & 0.529 & 0.325 & 0.403 & 0.226 & 0.302 & 0.146 & 0.503\\
% \textbf{TOAT-6-2} & 0.467 & 0.289 & 0.382 & 0.204 & 0.316 & 0.175 & 0.452\\
% \textbf{TOAT-6-lr} & 0. & 0. & 0. & 0.181 & 0.334 & 0.\\
% \textbf{TOAT-6-lr-s} & 0.468 & 0. & 0. & 0.207 & 0.322 & 0.\\
% \midrule
\textbf{OWAT} & 0.518 & \textbf{0.327} & \textbf{0.434} & \textbf{0.229} & \textbf{0.372} & \bf{0.203} & \bf{0.504}\\
% \textbf{OWAT-gamma03-30-6} & 0. & 0. & 0. & 0.175 & 0.345 & 0.\\
% \textbf{OWAT-gamma03-48-6} & 0. & 0. & 0. & 0.193 & 0.349 & 0.\\
%\midrule
\textbf{CWAT} & 0.513 & 0.325 & 0.433 & 0.224 & 0.367 & 0.199 & 0.503\\
% \textbf{CWAT-gamma03-30-6} & 0. & 0. & 0. & 0.196 &  & 0.\\
% \textbf{CWAT-gamma03-48-6} & 0. & 0. & 0. & 0.194 & 0.337 & 0.337.\\
% \textbf{CWAT-gamma03-48-10} & 0. & 0. & 0. & 0.194 & 0.336 & 0.337.\\
% \textbf{CWAT-10} & 0. & 0. & 0. & 0.191 & 0.331 & 0.\\
% \textbf{OWAT-s} & 0. & 0. & 0. & 0.215 & 0.299 & 0.\\
% \midrule
% \textbf{MIX-CWAT} & 0.488 & \textbf{0.377} & \textbf{0.440} & 
% \textbf{0.270} & 0.342 & 0.193 & 0.472\\
\bottomrule
\end{tabular}}%
% \end{center}%
\caption[]{The evaluation results of various adversarial trained SSD model with the VGG16-BN backbone network under FGSM PGD-10 attacks with $\epsilon=8$, CWA, and DAG in PASCAL VOC 2007 test set.}
\label{table:summaryforfree}
% \vspace{-0.2in}
\end{table}

\begin{table}
% \resizebox{1\columnwidth}{!}{%
% \begin{center}
\scalebox{0.85}{
\begin{tabular}{lcp{0.05\textwidth}p{0.05\textwidth}p{0.05\textwidth}p{0.05\textwidth}c}
\toprule
          \multicolumn{1}{l}{\multirow{2}{*}{\textbf{attack}}} & \multirow{2}{*}{\textbf{clean}}   & 
         \multicolumn{2}{c}{FGSM}   &
         \multicolumn{2}{c}{PGD-10} & \multirow{2}{*}{\textbf{CWA}} \\\cmidrule(lr){3-4}\cmidrule(lr){5-6}
         &&\textbf{$A_{cls}$}&\textbf{$A_{reg}$}&\textbf{$A_{cls}$}&\textbf{$A_{reg}$}\\
\midrule
\textbf{STD} & \textbf{0.451} & 0.133 & 0.167 & 0.030 & 0.029 & 0.003\\
% \midrule
% \textbf{MTD} & 0.178 & 0.103 & 0.114 & 0.076 & 0.080 & 0.035\\
\textbf{MTD}{\textsuperscript{1}} & 0.190 & 0.127 & 0.146 & 0.110 & 0.135 & 0.082\\
\textbf{MTD-fast} & 0.242 & 0.167 & 0.182 & 0.130 & 0.134 & 0.077\\
\midrule
\textbf{TOAT-6} & 0.182 & 0.120 & 0.148 & 0.098 & 0.123 & 0.074\\
\textbf{OWAT} & 0.211 & 0.129 & 0.169 & 0.100 & 0.140 & 0.074\\
\textbf{CWAT} & 0.237 & \textbf{0.168} & \textbf{0.189} & \textbf{0.142} & \textbf{0.155} & \textbf{0.092}\\
\bottomrule

\end{tabular}%
% \end{center}%
}
\caption[]{The adversarial robustness of each model using SSD VGG16-BN model under FGSM, PGD-10, and CWA attacks with $\epsilon=8$ in the MS-COCO test set.}
\label{table:cocoforfree}
% \vspace{-0.2in}
\end{table}

\subsubsection{\bf{Attack under Different Number of PGD Steps and Different Budgets}}
To evaluate the performance of the proposed adversarial training for object detection and compare with previous methods, we first attack the models using the adversarial examples generated with different number of PGD steps. As shown in Figure~\ref{fig:diffstep}, the proposed OWAT and CWAT both can enhance the robustness for these settings. With the proposed CWAT, the performance can be significantly enhanced as compared with our implemented MTD-fast where MTD~\cite{DBLP:journals/corr/abs-1907-10310} is the recent state-of-the-art adversarial training method for object detection. In addition, by taking both training time and the training settings of other related works into consideration, we choose PGD-10 to generate the adversarial examples for training. Moreover, we also evaluate the robustness of models under different budgets as shown in Figure~\ref{fig:diffeps}. We also evaluate each model under the adversarial attacks with different budgets, and we summarize the results in Appendix.
% \subsubsection{\bf{Attack under Different Budgets}}
% To evaluate the robustness of models under different budgets, we visualize the detection results of standard SSD model trained with clean images. As shown in Figure~\ref{fig:diffeps}, we can find that the attack strength of the perturbation will be stronger with the larger budget. To compromise between the natural accuracy and the adversarial robustness, we choose the attack budget of $\epsilon=8$ to perform the adversarial training. We also evaluate each model under the adversarial attacks with different budgets, and we summarize the results in Figure~\ref{fig:difepsmodel} which shows the robustness of the proposed algorithms can significantly enhance the performance using different budgets.

\subsection{The Impact of Different Thresholds \texorpdfstring{$\beta_{cls}$}{} and \texorpdfstring{$\beta _{reg}$}{} in Multi-task Adversarial Training}
To analyze the effect of the ``clipping'' parameters of $\beta_{cls}$ and $\beta _{reg}$, we further explore the proposed TOAT in different clipping parameters as follows.

\begin{itemize}

\item {\bf \textit{TOAT-inf}}: the model trained using TOAT and with the values of $\beta _{cls}$ and $\beta _{reg}$ set as $\infty$.

\item {\bf \textit{TOAT-i}}: the model trained using TOAT and with the values of $\beta _{cls}$ and $\beta _{reg}$ as i.

\item {\bf \textit{TOAT-k-l}}: the model trained using TOAT and with the values of $\beta _{cls}$ as $k$ and $\beta _{reg}$ as $l$.

\end{itemize}

As shown in Table~\ref{table:summaryfortoat}, % we can find that although TOAT cannot get the highest performance than the proposed OWAT and CWAT, it can still enhance the robustness of SSD model with the proper clipping parameters under various attacks including FGSM, CWA and DAG\footnote{We implement DAG algorithm into our one-stage models, the results are similar with the UEA experiments.}. If we compare with MTD-fast, we can find that TOAT could result in the better performance. Moreover,
we can find that $\beta _{cls}$ and $\beta _{reg}$ have important impacts on the adversarial robustness. As a result, the ``clipping'' parameters can prevent one of the task-oriented losses from dominating by the other. To choose the proper clipping thresholds, we first run two experiments with same $\beta _{cls}$ and $\beta _{reg}$ called TOAT-i. For example, we set $\beta _{cls} = \beta _{reg} = 2$ and $\beta _{cls} = \beta _{reg} = 4$ to analyze the impact on different parameters. We can explore that TOAT-2 has better performance defended against $l_{reg}$ attacks. On the other hand, TOAT-4 has better performance defended against $l_{cls}$ attacks. Therefore, if $\beta _{cls}$ and $\beta _{reg}$ are in different settings and and balance both task-oriented losses, we can optimize the adversarial training and get the better robustness. Therefore, we test TOAT with different $\beta _{cls}$ and $\beta _{reg}$ called TOAT-k-l. After a simple testing process, we choose the clipping parameters $\beta _{cls} = \beta _{reg} = 6$ as the final set and implement it into the OWAT and CWAT models.

\subsection{The Impact of Task-oriented Attack and Class-wise Attack for Object Detection}
As described in the methodology section, not all of the objects in an image would be attacked successfully. To analyze the potential influences between each object in an image, we explore this impact by delving into the object losses with adversarial attack in each class and by comparing the clean SSD model with CWAT.
\begin{table}
% \resizebox{1\columnwidth}{!}{%
% \begin{center}
\scalebox{0.63}{
\begin{tabular}{lccccccccc}
\toprule
& \multicolumn{3}{c}{\bf{Clean SSD}}&
  \multicolumn{3}{c}{\bf{OWAT SSD}} &
  \multicolumn{3}{c}{\bf{CWAT SSD}}  \\\cmidrule(lr){2-4}\cmidrule(lr){5-7}\cmidrule(lr){8-10}
& MDA & 
OWA & 
CWA &
MDA & 
OWA & 
CWA &
MDA & 
OWA & 
CWA
\\
\midrule
\textbf{aeroplane} & 0.010 & 0.019 & 0.002 & 0.383 & 0.370 & 0.365 & 0.354 & 0.373 & 0.353\\
\textbf{bicycle} & 0.024 & 0.092 & 0.001 & 0.354 & 0.365 & 0.357 & 0.334 & 0.340 & 0.338\\
\textbf{bird} & 0.003 & 0.003 & 0.001 & 0.118 & 0.117 & 0.103 & 0.106 & 0.110 & 0.111\\
\textbf{boat} & 0.006 & 0.001 & 0.000 & 0.186 & 0.174 & 0.126 & 0.165 & 0.161 & 0.116\\
\textbf{bottle} & 0.037 & 0.019 & 0.001 & 0.101 & 0.101 & 0.099 & 0.101 & 0.102 & 0.100\\
\textbf{bus} & 0.023 & 0.011 & 0.001 & 0.292 & 0.316 & 0.278 & 0.298 & 0.312 & 0.266\\
\textbf{car} & 0.020 & 0.010 & 0.092 & 0.409 & 0.423 & 0.425 & 0.411 & 0.429 & 0.435\\
\textbf{cat} & 0.000 & 0.001 & 0.000 & 0.148 & 0.162 & 0.100 & 0.146 & 0.157 & 0.087\\
\textbf{chair} & 0.005 & 0.004 & 0.001 & 0.095 & 0.095 & 0.109 & 0.109 & 0.107 & 0.109\\
\textbf{cow} & 0.000 & 0.002 & 0.000 & 0.054 & 0.050 & 0.058 & 0.089 & 0.069 & 0.033\\
\textbf{diningtable} & 0.005 & 0.005 & 0.000 & 0.317 & 0.317 & 0.220 & 0.288 & 0.290 & 0.207\\
\textbf{dog} & 0.003 & 0.000 & 0.000 & 0.080 & 0.083 & 0.071 & 0.116 & 0.125 & 0.093\\
\textbf{horse} & 0.007 & 0.003 & 0.002 & 0.369 & 0.373 & 0.277 & 0.336 & 0.329 & 0.284\\
\textbf{motorbike} & 0.002 & 0.003 & 0.002 & 0.317 & 0.328 & 0.277 & 0.301 & 0.317 & 0.265\\
\textbf{person} & 0.035 & 0.012 & 0.016 & 0.317 & 0.317 & 0.311 & 0.314 & 0.319 & 0.316\\
\textbf{pottedplant} & 0.007 & 0.091 & 0.000 & 0.099 & 0.097 & 0.095 & 0.096 & 0.096 & 0.094\\
\textbf{sheep} & 0.001 & 0.012 & 0.001 & 0.144 & 0.150 & 0.161 & 0.173 & 0.156 & 0.149\\
\textbf{sofa} & 0.001 & 0.000 & 0.000 & 0.155 & 0.174 & 0.134 & 0.156 & 0.160 & 0.142\\
\textbf{train} & 0.018 & 0.002 & 0.001 & 0.315 & 0.321 & 0.259 & 0.251 & 0.262 & 0.248\\
\textbf{tvmonitor} & 0.007 & 0.004 & 0.002 & 0.249 & 0.251 & 0.241 & 0.253 & 0.264 & 0.235\\
\midrule
\textbf{mAP} & 0.011 & 0.015 & \bf{0.006} & 0.225 & 0.229 & \bf{0.203} & 0.220 & 0.224 & \bf{0.199}\\ 
\bottomrule
\end{tabular}%
% \end{center}%
}
\caption[]{The average precision of each category under different PGD-10 attacks including MDA, OWA, and OWA with different robust SSD models in PASCAL VOC 2007 test set.}
\label{table:apincategory}
\vspace{-0.2in}
\end{table}
% \subsubsection{\textit{The Robustness of Different SSD Models on Each Category}}
\\ \\
\textbf{The Robustness of Different SSD Models on Each Category: }To verify the effectiveness of the proposed approaches and to improve the adversarial robustness for each object class evenly, we show the per-class performances after various adversarial attacks. As shown in Table \ref{table:apincategory}, the proposed approach significantly improves the adversarial robustness for each class, which also proves the effectiveness of our approach.
\\ \\
% \subsubsection{\textit{The Effectiveness of Different Attack Methods on Each Category}}
\textbf{The Effectiveness of Different Attack Methods on Each Category:} To delve into the effectiveness of different proposed attack methods, we can also figure out by the per-class performances after defense these adversarial attacks. As shown in Table \ref{table:apincategory}, the MDA proposed by \cite{DBLP:journals/corr/abs-1907-10310} is similar to the proposed OWA in this paper. Additionally, the proposed CWA is much more powerful than others, which also proves our concept mentioned in
% Section~\ref{sec:intro}
the introduction section that the class-wise adversarial attack can attack each class in a balanced way. However, there are few categories OWAT and CWAT will fail such as bicycle and car. In our observation, the categories which will cause the attack fail almost present as a large object in the images. Due to the SSD limitation of the large scale detection, it may cause the attack fail toward the large objects. 

% \subsubsection{\textit{The Robustness of Different SSD Models on Different Object Number in an Image}}
% Note that CWAT is less effective when the number of classes in a single image is few.
% Therefore, we independently analyze the CWAT performance with the subset that contains more than one class in an image.
% The results are illustrated in Fig.
% It can be seen that CWAT outperforms MTD, especially for multi-class images.

\subsection{Evaluation using Different Network Architecture} We also evaluate the SSD models using different backbone networks. In addition to the original VGG-16BN backbone, we also use the SSD model with the ResNet-50 as a backbone from the implementation of ScratchDet \cite{DBLP:journals/corr/abs-1810-08425}. 
As shown in Table~\ref{table:diffmodels}, the proposed method can improve the robustness of the SSD models significantly by 20\% to 30\% mAP across different backbone networks, demonstrating that the proposed CWAT can consistently improve the adversarial robustness of the object detectors with different backbone networks.
% Furthermore, to test the transferability of the CWAT, we also evaluate on two one-stage object detection model.
% They are RetinaNet \cite{DBLP:journals/corr/abs-1708-02002} and Yolo v3 \cite{DBLP:journals/corr/abs-1804-02767}. 
% For a fair comparison, we adjust the number of epochs so that the adversarial training time of RetinaNet and Yolo v3 is almost the same as SSD.
% The results are also shown in Table~\ref{table:diffmodels}.
% The CWAT outperforms MTD.

%
\begin{table}
% \resizebox{1\columnwidth}{!}{%
% \begin{center}
\centering
\scalebox{0.66}{
% \begin{tabular}{llcp{0.03\textwidth}p{0.03\textwidth}p{0.03\textwidth}p{0.03\textwidth}cc}
\begin{tabular}{llccccccc}
\toprule
           \multicolumn{2}{c}{\multirow{2}*{\textbf{attack}}} & \multirow{2}*{\textbf{clean}}   & 
         \multicolumn{2}{c}{FGSM}   &
         \multicolumn{2}{c}{PGD-10}  & \multirow{2}*{\textbf{CWA}} &
         \multirow{2}*{DAG}
         \\\cmidrule(lr){4-5}\cmidrule(lr){6-7}
         \multicolumn{1}{c}{}&&&\textbf{$A_{cls}$}&\textbf{$A_{reg}$}&\textbf{$A_{cls}$}&\textbf{$A_{reg}$} & \\ 
\midrule
 \multirow{2}*{\textbf{VGG16}} &\textbf{STD} & 0.752 & 0.162 & 0.250 & 0.012 & 0.043 & 0.006 & 0.291\\
 & \textbf{CWAT} & 0.513 & \bf{0.324} & \bf{0.433} & \bf{0.222} & \bf{0.366} & \bf{0.199} & 0.503
\\
\midrule
% \cmidrule{2-10}
 \multirow{2}*{\textbf{ResNet50}} &\textbf{STD} & 0.806 & \bf{0.344} & \bf{0.420} & 0.070 & 0.062 & 0.016 & 0.795\\
 & \textbf{CWAT} & 0.483 & 0.318 & 0.401 & \bf{0.244} & \bf{0.340} & \bf{0.250} & 0.482\\
%\midrule
% \multirow{2}*{\textbf{RetinaNet}} & \multirow{2}*{\textbf{ResNet50}} & \textbf{STD} & 0.759 & 0.223 & 0.399 & 0.013 & 0.021 & 0.033 & 0.210\\
% && \textbf{CWAT} & 0.562 & \textbf{0.321} & \textbf{0.475} & \textbf{0.186} & \textbf{0.352} & \textbf{0.226} & 0.561\\
\bottomrule
\end{tabular}%
% \end{center}%
}
\caption[]{The robustness of different one-stage models including SSD with different backbone networks under FGSM ,PGD-10, and CWA attacks with $\epsilon=8$ in the PASCAL VOC 2007 test set.}
\label{table:diffmodels}
% \vspace{-0.2in}
\end{table}
\begin{table}
% \resizebox{1\columnwidth}{!}{%
% \begin{center}
\centering
\scalebox{0.75}{
\begin{tabular}{lcp{0.05\textwidth}p{0.05\textwidth}p{0.05\textwidth}p{0.05\textwidth}cc}
\toprule
           \multirow{2}{*}{\textbf{attack}} & \multirow{2}{*}{\textbf{clean}}   & 
         \multicolumn{2}{c}{FGSM}   &
         \multicolumn{2}{c}{PGD-10} & 
         \multirow{2}{*}{\textbf{CWA}} &
         \multirow{2}{*}{DAG} \\\cmidrule(lr){3-4}\cmidrule(lr){5-6}
         &&\textbf{$A_{cls}$}&\textbf{$A_{reg}$}&\textbf{$A_{cls}$}&\textbf{$A_{reg}$}\\
\midrule
\textbf{TOAT-inf} & 0.469 & 0.266 & 0.374 & 0.173 & 0.310 & 0.165 & 0.456\\
\textbf{TOAT-2} & 0.535 & 0.302 & 0.408 & 0.191 & 0.297 & 0.132 & 0.502\\
\textbf{TOAT-4} & 0.477 & 0.288 & 0.396 & 0.190 & 0.327 & 0.183 & 0.463\\
\textbf{TOAT-6} & 0.430 & 0.300 & 0.397 & {0.218} & {0.334} & {0.197} & 0.466\\
\textbf{TOAT-8} & 0.343 & 0.289 & 0.389 & 0.204 & 0.331 & 0.189 & 0.457\\
\textbf{TOAT-2-4} & 0.512 & 0.304 & 0.388 & 0.211 & 0.292 & 0.131  & 0.487\\
\textbf{TOAT-4-2} & 0.479 & 0.296 & 0.404 & 0.203 & {0.337} & 0.178 & 0.472\\
\textbf{TOAT-2-6} & 0.529 & 0.325 & 0.403 & {0.226} & 0.302 & 0.146 & 0.503\\
\textbf{TOAT-6-2} & 0.467 & 0.289 & 0.382 & 0.204 & 0.316 & 0.175 & 0.452\\
\bottomrule
\end{tabular}}%
% \end{center}%
\caption[]{The evaluation results of different clipping hyperparameters of the proposed TOAT for the classification and regression losses for the adversarially trained SSD model with the VGG16-BN backbone network under FGSM PGD-10 attacks with $\epsilon=8$, CWA, and DAG in PASCAL VOC 2007 test set.}
\label{table:summaryfortoat}
% \vspace{-0.2in}
\end{table}

\section{Conclusion}\label{sec:conclusion}
In this work, we present several robust adversarial training for object detection. For a given image, the proposed approach can generate an effective universal adversarial perturbation to simultaneously attack all the occurred objects in the image through jointly maximizing the respective loss for each object. Additionally, the proposed class-wise adversarial training for object detection can not only balances the influence of each class but also effectively and evenly improves adversarial robustness of trained models for all the object classes as compared with the previous defense methods. Meanwhile, with the recent development of fast adversarial training, we provide a fast version of the proposed algorithm, which can be trained faster than the traditional adversarial training while keeping performance comparable. 
With extensive experiments on the challenging PASCAL-VOC and MS-COCO datasets, the evaluation results demonstrate that the proposed defense methods can effectively enhance the robustness of the object detection models.
\\
\\
\noindent\textbf{Acknowledgement}
This work is supported by Ministry of Science and Technology (MOST), Taiwan (R.O.C.), under Grants No. 108-2218-E-001-004-MY2, 109-2221-E-001-020-, and 109-2221-E-001-016-.

{\small
\bibliographystyle{ieee_fullname}
\bibliography{egbib}
}

\clearpage

% ==============================================================================================

\subsection*{A. Attack under Different Number of PGD Steps and Different Budgets}
\begin{figure}[ht]
    \begin{center}
  \subfloat[$A_{cls}$ PGD-10 attack\label{PGD_10_3a}]{%
       \includegraphics[width=0.5\linewidth]{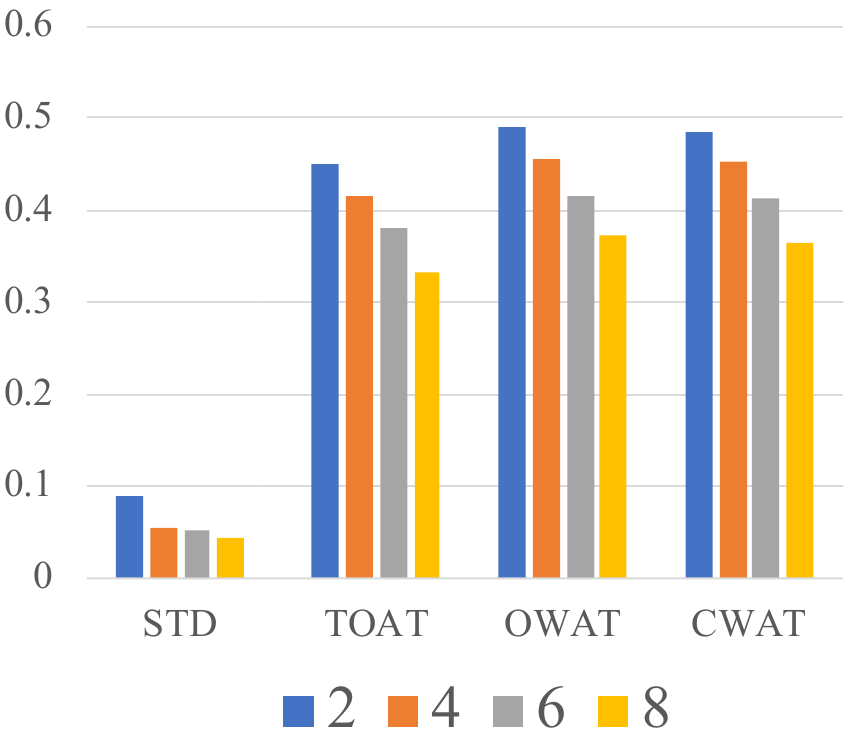}}
    \hfill
  \subfloat[$A_{reg}$ PGD-10 attack\label{PGD_10_3b}]{%
        \includegraphics[width=0.48\linewidth]{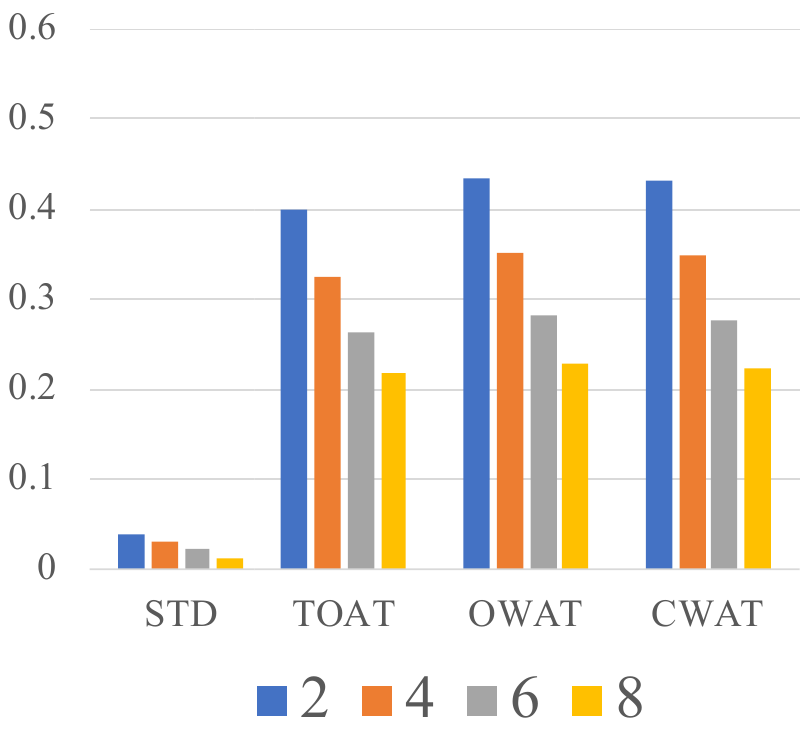}}
    \hfill
  \caption{The robustness of each model under PGD-10 attacks from different budgets in PASCAL VOC 2007 test set.}
  \label{fig:difepsmodel}
  \end{center}
\end{figure}

To evaluate the performance of the proposed adversarial training for object detection and compare with previous methods, we first attack the models using the adversarial examples generated with different number of PGD steps. As shown in Figure~\ref{fig:Adversarialimage}, the proposed OWAT and CWAT both can enhance the robustness for these settings. With the proposed CWAT, the performance can be significantly enhanced as compared with our implemented MTD-fast where MTD~\cite{DBLP:journals/corr/abs-1907-10310} is the recent state-of-the-art adversarial training method for object detection. In addition, by taking both training time and the training settings of other related works into consideration, we choose PGD-10 to generate the adversarial examples for training. Moreover, we also evaluate each model under the adversarial attacks with different budgets as shown in Figure~\ref{fig:difepsmodel} and Figure~\ref{fig:diffeps}.

\begin{figure*}[t]
    \begin{center}
  \subfloat[$\epsilon=0$\label{1a}]{%
       \includegraphics[width=0.18\linewidth]{LaTeX/ori-img.jpg}}
       \hfill
  \subfloat[$\epsilon=2$\label{1b}]{%
       \includegraphics[width=0.18\linewidth]{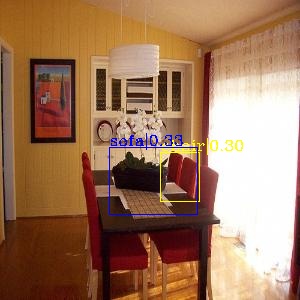}}
       \hfill
  \subfloat[$\epsilon=4$\label{1c}]{%
       \includegraphics[width=0.18\linewidth]{LaTeX/adv-4.jpg}}
       \hfill
  \subfloat[$\epsilon=6$\label{1d}]{%
       \includegraphics[width=0.18\linewidth]{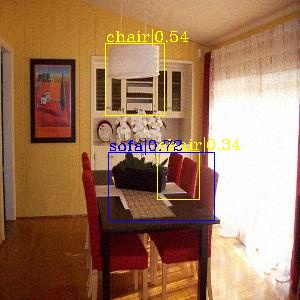}}
       \hfill
  \subfloat[$\epsilon=8$\label{1e}]{%
       \includegraphics[width=0.18\linewidth]{LaTeX/adv-8.jpg}}
  \caption{The detection results of the class-wise adversarial attack with PGD-10 in different $\epsilon$s, using the clean SSD as the targeted model. White label, yellow label, magenta label, blue label, green label, and black label represent classes of the dining table, chair, potted plant, sofa, tv-monitor, and respectively.}
  \label{fig:diffeps}
  \end{center}
\end{figure*}

\subsection*{B. The Impact of Fast Adversarial Training}
It can be 7 to 30 times faster than the corresponding PGD-based adversarial training as mentioned in \cite{DBLP:journals/corr/abs-1808-07945}. Moreover, the original PGD-10 adversarial training in \cite{DBLP:journals/corr/abs-1907-10310} needs 23 back-propagations (each task costs 10 to generate task-oriented adversarial example, 2 to determine which example is used for final training, and 1 back-propagation for the final model update) per-iteration. On the other hand, our proposed methods only uses 2 back-propagation (1 CWT, and 1 for the update). Therefore, the original approach MTD would take additional 21 back-propagations. For the experiments, our fast CWAT is 3.19x faster than MTD with 4 2080Ti GPUs and batch size 14 for each GPU.

\subsection*{C. More Details for Training}
For the proposed adversarial training, we select all the positive anchors after each anchor has predicted. 
The positive anchors in the SSD are those that their IOUs between the ground truth are greater than 0.5. 
When we calculate the loss, we use all positive anchors and choose a certain percentage of negative anchors. 
Then we utilize this loss to calculate the attack gradient.
Note that this procedure does not include non-maximum suppression (NMS).
The same as the original SSD training, we do not use NMS when training, and the NMS is used in inference and test.
The proposed method will attack all positive anchors rather than the single anchor that has the maximum IOU.

\subsection*{D. The Results under Different Kinds of Attacks}
\begin{figure*}[t]
% \vspace{-0.05in}
    \begin{center}
  \subfloat[Clean Image Result\label{fig_1a}]{%
       \includegraphics[width=0.18\linewidth]{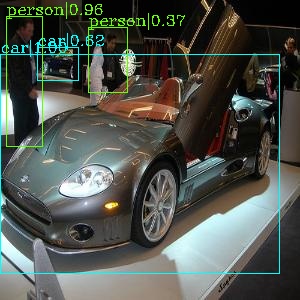}}
    \hfill
  \subfloat[Vanilla Adversarial Attack\label{fig_1b}]{%
       \includegraphics[width=0.18\linewidth]{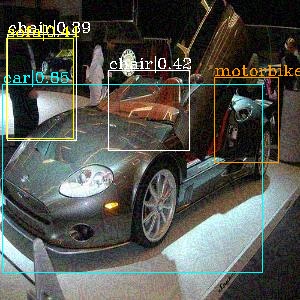}}
    \hfill
  \subfloat[Multi-task domain attack\label{fig_1c}]{%
       \includegraphics[width=0.18\linewidth]{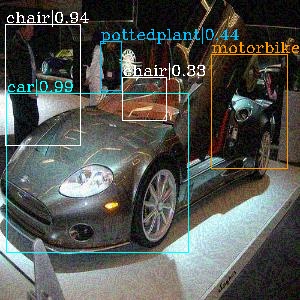}}
    \hfill
  \subfloat[Object-wise attack\label{fig_1d}]{%
        \includegraphics[width=0.18\linewidth]{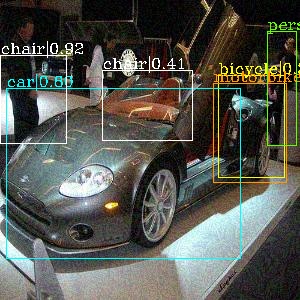}}
    \hfill
  \subfloat[Class-wise Attack\label{fig_1e}]{%
        \includegraphics[width=0.18\linewidth]{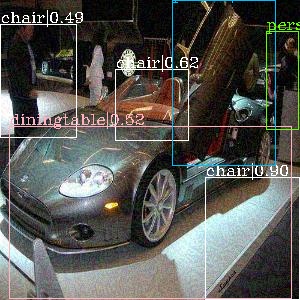}}
    \hfill
  \caption{Detection results after attacked by different adversarial examples to the vanilla SSD model. (a) the detection result of a clean image, (b) the detection result after attacked by the adversarial example crafted through the 20-step PGD optimization with the budget $\epsilon=16$ on the multi-task loss as described in equation 1, (c) the detection result after the multi-task domain attack which we follow \cite{DBLP:journals/corr/abs-1907-10310} to implement it, (d) the detection result after the proposed object-wise attack, (e) the detection result after the proposed class-wise attack.}
  \label{fig:Adversarialimage}
  \end{center}
\end{figure*}

The visualization of the detection results of an image under different attack are shown in Figure~\ref{fig:Adversarialimage}. These detection examples show the adversarial examples generated by the proposed method can more evenly attack all the objects occurred in the image than \ref{fig_1b} and \ref{fig_1c} which use total losses to generate the adversarial attack.

\subsection*{E. More Qualitative Results for the Proposed CWAT Detector}
\begin{figure*}[t]
  \subfloat[\centering No attack; Model: \textbf{STD}]{%
       \includegraphics[width=0.23\linewidth]{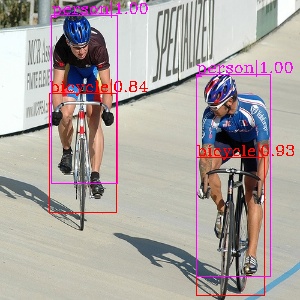}}
    \hspace{0.2cm}
  \subfloat[\centering Attack: \textbf{CWA}; Model: \textbf{STD}]{%
       \includegraphics[width=0.23\linewidth]{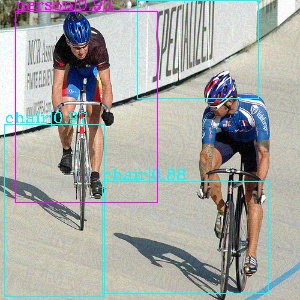}}
   \hspace{0.2cm}
  \subfloat[\centering Attack: \textbf{CWA}; Model: \textbf{CWAT}]{%
       \includegraphics[width=0.23\linewidth]{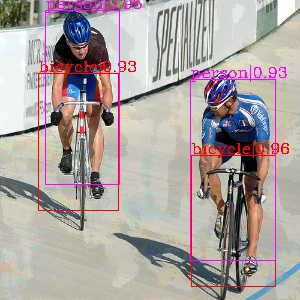}}
    \hspace{0.2cm}
  \subfloat[\centering Attack: \textbf{DAG}; Model: \textbf{CWAT}]{%
        \includegraphics[width=0.23\linewidth]{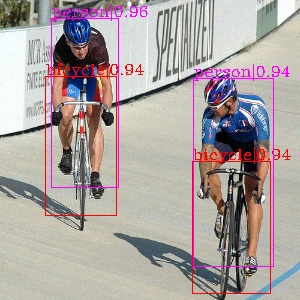}}
        \\
  \subfloat[\centering No attack; Model: \textbf{STD}]{%
       \includegraphics[width=0.23\linewidth]{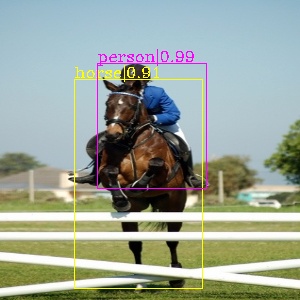}}
    \hspace{0.2cm}
  \subfloat[\centering Attack: \textbf{CWA}; Model: \textbf{STD}]{%
       \includegraphics[width=0.23\linewidth]{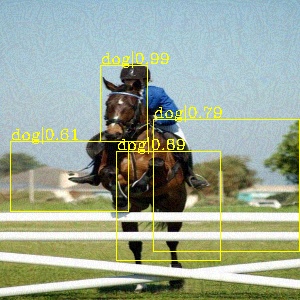}}
   \hspace{0.2cm}
  \subfloat[\centering Attack: \textbf{CWA}; Model: \textbf{CWAT}]{%
       \includegraphics[width=0.23\linewidth]{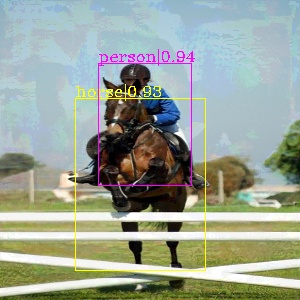}}
    \hspace{0.2cm}
  \subfloat[\centering Attack: \textbf{DAG}; Model: \textbf{CWAT}]{%
        \includegraphics[width=0.23\linewidth]{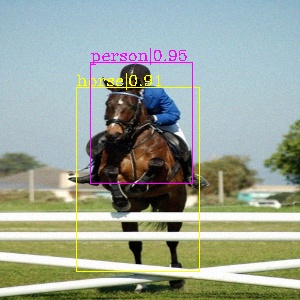}}
        \\
  \subfloat[\centering No attack; Model: \textbf{STD}]{%
       \includegraphics[width=0.23\linewidth]{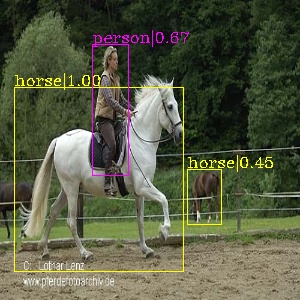}}
    \hspace{0.2cm}
  \subfloat[\centering Attack: \textbf{CWA}; Model: \textbf{STD}]{%
       \includegraphics[width=0.23\linewidth]{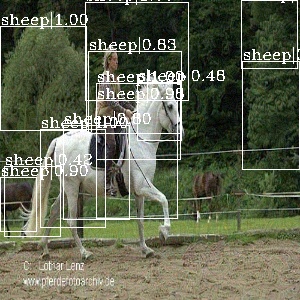}}
   \hspace{0.2cm}
  \subfloat[\centering Attack: \textbf{CWA}; Model: \textbf{CWAT}]{%
       \includegraphics[width=0.23\linewidth]{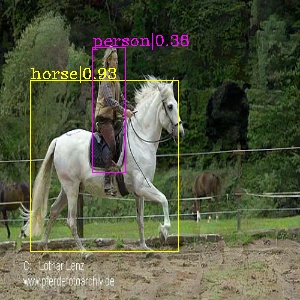}}
    \hspace{0.2cm}
  \subfloat[\centering Attack: \textbf{DAG}; Model: \textbf{CWAT}]{%
        \includegraphics[width=0.23\linewidth]{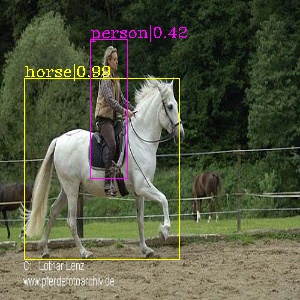}}
        \\
  \subfloat[\centering No attack; Model: \textbf{STD}]{%
       \includegraphics[width=0.23\linewidth]{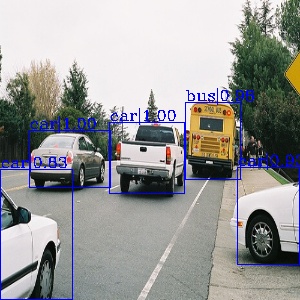}}
    \hspace{0.2cm}
  \subfloat[\centering Attack: \textbf{CWA}; Model: \textbf{STD}]{%
       \includegraphics[width=0.23\linewidth]{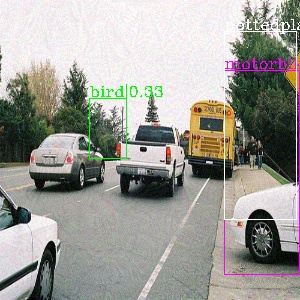}}
   \hspace{0.2cm}
  \subfloat[\centering Attack: \textbf{CWA}; Model: \textbf{CWAT}]{%
       \includegraphics[width=0.23\linewidth]{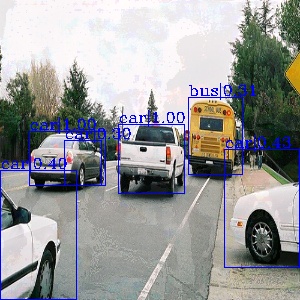}}
    \hspace{0.2cm}
  \subfloat[\centering Attack: \textbf{DAG}; Model: \textbf{CWAT}]{%
        \includegraphics[width=0.23\linewidth]{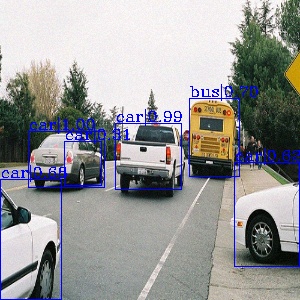}}
  \caption{Visualization results (attack budget = 8/255). The first column is STD model with no attack. The second column is STD model under class-wise attacks. The third column is CWAT defense model against class-wise attacks. The fourth column is CWAT defense model against DAG\cite{xie2017adversarial} attack.}
  \label{fig:AdversarialimageExample}
%   \end{center}
% \vspace{-0.25in}
\end{figure*}

Figure~\ref{fig:AdversarialimageExample} illustrates the visualization results of object detection for the standard and the proposed CWAT models under different adversarial attacks for object detection.
The first column is the detection results of the standard model (STD) upon clean images.
The second column is the detection results of the standard model under the proposed class-wise attack (CWA). 
As the figure shown, all the objects in the images are detected incorrectly.
The CWA is effective to fool the object detection model as demonstrated in the main paper.
Furthermore, the third and fourth columns are the detection results of the CWAT model to defend against CWA and DAG attacks \cite{xie2017adversarial}.
As the figures illustrated, the detection results using the proposed CWAT trained detector are almost the same as the ones using the clean model upon clean images. This further confirms the effectiveness of the proposed CWAT method.

\end{document}